\documentclass[runningheads]{llncs}
\usepackage{hyperref}
\usepackage[title]{appendix}
\usepackage{academicons}
\usepackage[utf8]{inputenc}
\usepackage{amsmath}
\usepackage{amssymb}
\usepackage{csquotes}
\usepackage[english]{babel}
\usepackage{scrextend}
\usepackage{wrapfig}
\usepackage{algorithm}
\usepackage{multicol}
\usepackage[noend]{algpseudocode}
\usepackage{mathabx}
\usepackage{adjustbox}

\algrenewcommand\algorithmicforall{\textbf{foreach}}
\algrenewcommand\algorithmicindent{.8em}

\usepackage{hyperref}
\usepackage[title]{appendix}
\usepackage{academicons}
\usepackage{xcolor}
\usepackage{caption}
\usepackage{wasysym}
\usepackage{enumitem}
\usepackage{color}
\usepackage{forloop}
\usepackage{ifthen}
\usepackage{graphicx}
\usepackage{{scrlayer-scrpage}}
\usepackage{cite}
\ihead{Learning to Sparsify TSP Instances}
\ohead{\pagemark}
\usepackage{scalerel}
\usepackage{tikz}
\usetikzlibrary{svg.path}
\usepackage[utf8]{inputenc}
\usepackage{array}
\usepackage[noend]{algpseudocode}
\newcolumntype{x}[1]{>{\centering\let\newline\\\arraybackslash\hspace{0pt}}p{#1}}
\algrenewcommand\algorithmicforall{\textbf{foreach}}
\algrenewcommand\algorithmicindent{.8em}

\newcounter{qr}

\newcounter{ql}

\newlength{\qt}

\newcounter{itemnummer}
\newcommand{\Qitem}[2][]{
\ifthenelse{\equal{#1}{}}{\stepcounter{itemnummer}}{}
\ifthenelse{\equal{#1}{a}}{\stepcounter{itemnummer}}{}
\begin{enumerate}[topsep=2pt,leftmargin=2.8em]
\item[\textbf{\arabic{itemnummer}#1.}] #2
\end{enumerate}
}

\definecolor{bgodd}{rgb}{0.8,0.8,0.8}
\definecolor{bgeven}{rgb}{0.9,0.9,0.9}
\newcounter{itemoddeven}
\newlength{\gb}
\newcommand{\QItem}[2][]{
\setlength{\gb}{\linewidth}
\addtolength{\gb}{-5.25pt}
\ifthenelse{\equal{\value{itemoddeven}}{0}}{%
\noindent\colorbox{bgeven}{\hskip-3pt\begin{minipage}{\gb}\Qitem[#1]{#2}\end{minipage}}%
\stepcounter{itemoddeven}%
}{%
\noindent\colorbox{bgodd}{\hskip-3pt\begin{minipage}{\gb}\Qitem[#1]{#2}\end{minipage}}%
\setcounter{itemoddeven}{0}%
}
}
\RequirePackage{etoolbox}
\DeclareRobustCommand\orcidicon[1]{\href{https://orcid.org/#1}{\mbox{\scalerel*{
\begin{tikzpicture}[yscale=-1,transform shape]
\pic{orcidlogo};
\end{tikzpicture}
}{|}}}}

\definecolor{orcidlogocol}{HTML}{A6CE39}
\tikzset{
  orcidlogo/.pic={
    \fill[orcidlogocol] svg{M256,128c0,70.7-57.3,128-128,128C57.3,256,0,198.7,0,128C0,57.3,57.3,0,128,0C198.7,0,256,57.3,256,128z};
    \fill[white] svg{M86.3,186.2H70.9V79.1h15.4v48.4V186.2z}
                 svg{M108.9,79.1h41.6c39.6,0,57,28.3,57,53.6c0,27.5-21.5,53.6-56.8,53.6h-41.8V79.1z M124.3,172.4h24.5c34.9,0,42.9-26.5,42.9-39.7c0-21.5-13.7-39.7-43.7-39.7h-23.7V172.4z}
                 svg{M88.7,56.8c0,5.5-4.5,10.1-10.1,10.1c-5.6,0-10.1-4.6-10.1-10.1c0-5.6,4.5-10.1,10.1-10.1C84.2,46.7,88.7,51.3,88.7,56.8z};
  }
}

\begin{document}

\title{Learning to Sparsify Travelling Salesman Problem Instances}

\author{James Fitzpatrick\inst{1}\orcidicon{0000-0002-9419-7005} \and Deepak Ajwani\inst{2}\orcidicon{0000-0001-7269-4150} \and Paula Carroll\inst{1}\orcidicon{0000-0003-1029-1668}}

\authorrunning{Fitzpatrick et al.}

\institute{School of Business, University College Dublin \email{\{james.fitzpatrick1@ucdconnect.ie, paula.carroll@ucd.ie\}}\\ \and School of Computer Science, University College Dublin\\ \email{deepak.ajwani@ucd.ie}}

\maketitle             

\begin{abstract}
In order to deal with the high development time of exact and approximation algorithms for NP-hard combinatorial optimisation problems and the high running time of exact solvers, deep learning techniques have been used in recent years as an end-to-end approach to find solutions. However, there are issues of representation, generalisation, complex architectures, interpretability of models for mathematical analysis etc. using deep learning techniques. As a compromise, machine learning can be used to improve the run time performance of exact algorithms in a matheuristics framework. 
In this paper, we use a pruning heuristic leveraging machine learning as a pre-processing step followed by an exact Integer Programming approach. We apply this approach to sparsify instances of the classical travelling salesman problem.  Our approach learns which edges in the underlying graph are unlikely to belong to an optimal solution and removes them, thus sparsifying the graph and significantly reducing the number of decision variables. We use carefully selected features derived from linear programming relaxation, cutting planes exploration, minimum-weight spanning tree heuristics and various other local and statistical analysis of the graph. Our learning approach requires very little training data and is amenable to mathematical analysis. 
We demonstrate that our approach can reliably prune a large fraction of the variables in TSP instances from TSPLIB/MATILDA ($>85\%$) while preserving most of the optimal tour edges. Our approach can successfully prune problem instances even if they lie outside the training distribution, resulting in small optimality gaps between the pruned and original problems in most cases. Using our learning technique, we discover novel heuristics for sparsifying TSP instances, that may be of independent interest for variants of the vehicle routing problem.

\keywords{Travelling Salesman Problem  \and Graph Sparsification \and Machine Learning \and Linear Programming \and Integer Programming}
\end{abstract}
\section{Introduction}
\label{sec:intro}
Owing to the high running time of exact solvers on many instances of NP-hard combinatorial optimisation problems (COPs), there has been a lot of research interest in leveraging machine learning techniques to speed up the computation of optimisation solutions. In recent years, deep learning techniques have been used as an end-to-end approach (see e.g.,~\cite{vinyals2015pointer}) for efficiently solving COPs. However, these approaches generally suffer from (i) limited generalisation to larger size problem instances and limited generalisation from instances of one domain to another domain, (ii) need for increasingly complex architectures to improve generalisability and (iii) inherent black-box nature of deep learning that comes in the way of mathematical analysis~\cite{DiCaro}. In particular, the lack of interpretability of these models means that (1) we do not know which properties of the input instances are being leveraged by the deep-learning solver and (2) we cannot be sure that the model will still work as new constraints are required, which is typical in industry use-cases.

\hspace{1em}

\noindent
In contrast to the end-to-end deep learning techniques, there has been recent work (see e.g.~\cite{GGKM0B20}) to use machine learning as a component to speed-up or scale-up the exact solvers. In particular, Lauri and Dutta~\cite{LD19} recently proposed a framework to use machine learning as a pre-processing step to sparsify the maximum clique enumeration instances and scale-up the exact algorithms in this way. In this work, we build upon this framework and show that  integrating features derived from operations research and approximation algorithms into the learning component for sparsification can result in reliably pruning a large fraction of the variables in the classical Travelling Salesman Problem (TSP). Specifically, we use carefully selected features derived from linear programming relaxation, cutting planes exploration, minimum-weight spanning tree (MST) heuristics and various other local and statistical analysis of the graph to sparsify the TSP instances. With these features, we are able to prune more than 85\% of the edges on TSP instances from TSPLIB/MATILDA, while preserving most of the optimal tour edges. Our approach can successfully prune problem instances even if they lie outside the training distribution, resulting in small optimality gaps between the pruned and original problems in most cases. Using our learning technique, we discover novel heuristics for sparsifying TSP instances, that may be of independent interest for variants of the vehicle routing problem.

\hspace{1em}

\noindent
Overall, our approach consists of using a pruning heuristic leveraging machine learning (ML) as a pre-processing step to sparsify instances of the TSP, followed by an exact Integer Programming (IP) approach. We  learn which edges in the underlying graph are unlikely to belong to an optimal TSP tour and remove them, thus sparsifying the graph and significantly reducing the number of decision variables. Our learning approach requires very little training data, which is a crucial requirement for learning techniques dealing with NP-hard problems. The usage of well analysed intuitive features and more interpretable learning models means that our approach is amenable to mathematical analysis. For instance, by inserting the edges from the double-tree approximation in our sparsified instances, we can guarantee the same bounds on the optimality gap as a double-tree approximation. We hypothesise that our approach, integrating features derived from operations research and approximation algorithms into a learning component for sparsification, is likely to be useful in a range of COPs, including but not restricted to, vehicle routing problems.


\noindent
\paragraph{Outline.} The paper is structured as follows: Section~\ref{sec:relatedworks} describes the related works. In Section~\ref{sec:sparsification} we outline the proposed sparsification scheme: the feature generation, the sparsification model and post-processing techniques. Section~\ref{sec:experiments} contains the experimental setup, exposition on the computational experiments and results. Discussion and conclusions follow in Section~\ref{sec:discussion}. 

\section{Notation and Related Work}
\label{sec:relatedworks}
\vspace{-0.25cm}

Given a graph $G = (V, E; w)$ with a vertex set $V = \{1, ..., n\}$, an edge set $E = \{(u, v)| u, v \in V, u \neq v\}$ and a weighting function $w_{G}(e) \rightarrow \mathbb{Z}^{+}$ that assigns a weight to each edge $e \in E$, the goal of the TSP is to find a tour in $G$ that visits each vertex exactly once, starting and finishing at the same vertex,  with least cumulative weight. We denote by $m$ the number of edges $|E|$ and by $n$ the number of vertices $|V|$ of the problem.


\subsection{Exact, Heuristic and Approximate Approaches}
\vspace{-0.25cm}
The TSP is one of the most widely-studied COPs and has been for many decades; for this reason, many very effective techniques and solvers have been developed for solving them. Concorde is a well-known, effective exact solver which implements a branch and cut approach to solve a TSP IP
 \cite{applegate2009certification}, and has been used to solve very large problems. An extremely efficient implementation of the Lin-Kernighan heuristic is available at \cite{helsgaun2000effective}, which can find very close-to-optimal solutions in most cases. Approximation algorithms also exist for the metric TSP that permit the identification of solutions, with worst-case performance guarantees, in polynomial time \cite{christofides1976worst, serdyukov1978}. Many metaheuristic solution frameworks have also been proposed, using the principles of ant colony optimisation, genetic algorithms and simulated annealing among others \cite{dorigo1997ant, hopfield1985, braun1990solving}. In each of these traditional approaches to the TSP, there are lengthy development times and extensive problem-specific knowledge is required. If the constraints of the given problem are altered, the proposed solution method may no longer be satisfactory, possibly requiring further development.

\subsection{Learning to Solve Combinatorial Optimisation Problems}
\vspace{-0.25cm}


Inspired by the success of deep learning to solve natural language processing and computer vision tasks, the question of how effective similar techniques might be in COP solution frameworks arises. Interest in this research direction emerged following the work of Vinyals et al. \cite{vinyals2015pointer}, in which sequence-to-sequence neural networks were used as heuristics for small instances of three different COPs, including the TSP. A number of works quickly followed that solve larger problem instances and avoid the need for supervised learning where access to data is a bottleneck \cite{bello2016neural, nazari2018reinforcement, kool2018attention}. Graph neural networks \cite{scarselli2008graph} and transformer architectures \cite{vaswani2017attention} lead to significant speedups for learned heuristics, and have been demonstrated to obtain near-optimal solutions to yet larger TSP and vehicle routing problem (VRP) instances in seconds. Although they can produce competitive solutions to relatively small problems, these learning approaches appear to fail to generalise well to larger instance sizes, and most of these approaches require that the instance is Euclidean, encoding the coordinates of the vertices for feature computation. In cases of failure and poor solution quality, however, there is little possibility of interpreting why mistakes were made, making it difficult to rely on these models.

\subsection{Graph Sparsification}
\vspace{-0.25cm}

Graph sparsification is the process of pruning edges from a graph $G = (V, E)$ to form a subgraph $H = (V, E'\subset E)$ such that $H$ preserves or approximates some property of $G$ \cite{peleg1989, benczur1996approximate, fung2019general}. Effective sparsification is achieved if $|E'| \ll |E|$. The running time of a TSP solve can be reduced if it is possible to sparsify the underlying complete graph $K_{n}$ so that the edges of at least one optimal Hamiltonian cycle are preserved. The work of Hougardy and Schroeder \cite{hougardy2014edge} sparsifies the graph defining symmetric TSP instances exactly, removing a large fraction of the edges, known as ``useless" edges, that provably cannot exist in an optimal tour. Another heuristic approach due to Wang and Remmel \cite{wang2018method} sparsifies symmetric instances by making probabilistic arguments about the likelihood that an edge will belong to an optimal tour. Both of these approaches have proven successful, reducing computation time significantly for large instances, but are unlikely to be easy to modify for different problem variants. 


\hspace{1em}

\noindent
Recently, the sparsification problem has been posed as a learning problem, for which a binary classification model is trained to identify edges unlikely to belong to an optimal solution. Grassia et al. \cite{grassia2019learning} use supervised learning to prune edges that are unlikely to belong to maximum cliques in a multi-step sparsification process. This significantly reduces the computational effort required for the task. Sun et al. \cite{sun2020generalization} train a sparsifier for pruning edges from TSP instances that are unlikely to belong to the optimal tour. These approaches have the advantage that they can easily be modified for similar COP variants. The use of simpler, classical ML models lends them the benefits of partial interpretability and quick inference times. In the latter case however, it is assumed that a very large number of feasible TSP solutions can be sampled efficiently, which does not hold for all routing-type problems, and in neither case are guarantees provided that the resultant sparsified instance admits a feasible solution.

\section{Sparsification Scheme}
\label{sec:sparsification}
\vspace{-0.25cm}

The sparsification problem is posed as a binary classification task. Given some edge $e \in E$, we wish to assign it a label $0$ or $1$, where the label $0$ indicates that the associated edge variable should be pruned and the label $1$ indicates that it should be retained. We acquire labelled data for a set of graphs $\mathcal{G} = \{G_{1}, ..., G_{n}\}$ corresponding to TSP problem instances. For each graph $G_{i} = (V_{i}, E_{i})$ we compute the set of all $p_{i}$ optimal tours $\mathcal{T}_{i} = \{t^{1}_{i}, ..., t^{p_{i}}_{i}\}$ and for each edge $e \in E_{i}$ we compute a feature representation $\vec{q}_{e}$. Each tour $t_{i}$ has an implied set of edges $t_{i} \implies \epsilon_{i} \subset E_{i}$. Labelling each $e \in \bigcup_{j=1}^{p_{i}} \epsilon^{j}_{i} = \mathcal{E}_{i}$ with $1$ and each edge $e \in E_{i} \setminus \mathcal{E}_{i}$ as $0$, we train an edge classifier to prune edges that do not belong to the optimal tour. An optimal sparsifier would prune all but those edges belonging to optimal tours (potentially also solving the TSP). This classifier represents a binary-fixing heuristic in the context of an IP. In the following sections, we describe the feature representation that is computed for each edge and post-processing steps that are taken in order to make guarantees and to alleviate the effects of over-pruning.


\subsection{Linear Programming Features}
\vspace{-0.25cm}

We pose the TSP as an IP problem, using the DFJ formulation \cite{dantzig1954solution}. Taking $A_{ij}$ as the matrix of edge-weights, we formulate it as follows:

\begin{alignat}{3}
& \text{minimize} & z = \sum\limits_{i,j=1;i\neq j}^{n} A_{ij} x_{ij} & 
\label{tsp_obj} \\
& \text{subject to} \quad & \sum\limits_{i; i \neq j}^{n}  x_{ij} = 1, \qquad & j=1 ,..., n 
\label{tsp_deg_in} \\
&& \sum\limits_{j; j \neq i}^{n}  x_{ij} = 1, \qquad & i=1 ,..., n 
\label{tsp_deg_out} \\
&& \sum\limits_{(i,j) \in W} x_{ij} \leq |W| - 1, \qquad & W \subseteq V; \quad |W| \geq 3 
\label{tsp_sbt} \\
&& x_{ij} \in \{0,1\}, \qquad & i,j=1 ,..., m; \quad i \neq j 
\label{tsp_int} \\ 
\nonumber
\end{alignat}

\vspace{-1em}

\noindent
Useful information about the structure of a TSP problem can be extracted 
by inspecting solution vectors to linear relaxations of this IP, we can obtain insights into the candidacy of edges for the optimal tour. In fact, in several cases, for the MATILDA problem set, the solution to the linear relaxation at the root of the Branch and Bound (B\&B) tree is itself an optimal solution to the TSP. 

\hspace{1em}

\noindent
We denote the solution to the linear relaxation $z_{LP}$ of the integer programme at the root node of the B\&B tree by $\hat{\vec{x}}^{0}$. At this point, no variables have been branched on and  no subtour elimination cuts have been introduced. That is, the constraints (\ref{tsp_sbt}) are dropped and the constraints (\ref{tsp_int}) are relaxed as:

\begin{equation}
    x_{ij} \in [0,1] , \qquad \qquad i, j \in \{1, ..., m\}, i \neq j.
    \label{tsp_relax}
\end{equation}

One can strengthen this relaxation by introducing some subtour elimination constraints (\ref{tsp_sbt}) at the root node
. In this case, the problem to be solved remains a linear programming problem but several rounds of subtour elimination cuts are added. One can limit the computational effort expended in this regard by restricting the number of constraint-adding rounds with some upper bound $k$. The solution vector for this problem after $k$ rounds of cuts is denoted by $\tilde{\vec{x}}^{k}$. We can  also use as features the associated reduced costs $\vec{r}^{k}$, which are computed in the process of a Simplex solve, standardising their values as $\hat{\vec{r}}^{k}= \vec{r}^{k} / \max{\vec{r}^{k}}$. 

\hspace{1em}

\noindent
In order to capture broader information about the structure of the problem, stochasticity is introduced to the cutting planes approach. Inspired by the work of Fischetti and Monaci \cite{fischetti2014exploiting}, the objective of the problem is perturbed. Solving the perturbed problem results in different solution vectors, which can help us to explore the solution space. We solve the initial relaxation $z_{LP}$ in order to obtain a feasible solution, which can sometimes take a significant amount of computing effort. Subtour-elimination constraints are added to the problem for each subtour in the initial relaxation. Following this, $k$ copies of this problem are initialised. For each new problem, the objective coefficients $A_{ij}$ are perturbed, and the problem is re-solved. The normalised reduced costs are obtained from each perturbed problem and for each edge $(i,j)$ the mean reduced cost $\tilde{r}_{ij}$ is computed. We use the vector $\tilde{\vec{r}}$ of such values as a feature vector.

\subsection{Minimum Weight Spanning Tree Features}\label{ssec:num1}

The MST provides the basis for the Christofides–Serdyukov and double-tree approximation algorithms, which give  feasible solutions with optimality guarantees for a symmetric, metric TSP. Taking inspiration from these approximation algorithms, we use multiple MSTs to extract edges from the underlying graph that are likely to belong to the optimal solution, thereby allowing us to compute features using them.

\begin{wrapfigure}[12]{L}{0.55\textwidth}
\begin{minipage}{0.55\textwidth}
\vspace{-3em}
\begin{algorithm}[H]
\textbf{Input:} $G = (V, E)$
\begin{algorithmic}[1]
\caption{Minimum Spanning Tree Features \label{algo_mwst}}
    \State $H \leftarrow (V, E'' = \emptyset)$ 
    \For {$k \in \{1, 2, ...,  \lceil\log{n}\rceil\}$}
        \State $T = (V, E' \subset E) \leftarrow$ MST($G$)
        \State $G = (V, E) \leftarrow (V, E \setminus E')$
        \State $H = (V, E'') \leftarrow (V, E'' \cup E')$
        \State $w_{H}(e) = 1/k  \quad \forall e \in E'$
    \EndFor
\end{algorithmic}
\textbf{Output:} $H = (V, E'') $
\end{algorithm}
\end{minipage}
\end{wrapfigure}

First, a new graph $H = (V, \emptyset)$ is initialised with the vertex set but not the edge set of the complete graph $G = (V, E) = K_{n}$. For $k \ll n$ iterations the MST $T = (V, E')$ of $G$ is computed and the edges $E'$ are removed from $E$ and added to the edge set of $H$. Then, at each step, $ G(V, E)  \leftarrow G(V, E \setminus E')$, giving a new MST at each iteration with unique edges. The iteration at which edges are added to the graph $H$ is stored, so that a feature $\hat{q}^{k}_{ij} = 1_{i}/k$ may be computed, where $1_{i}$ is the indicator, taking unit value if the edge $e = (i, j) \in E$ was extracted at iteration $i$ and zero otherwise. 

\hspace{1em}

\noindent
Since the value of $k$ should be small, the vast majority of the edges will have zero-valued feature-values. This edge transferal mechanism can be used as a sparsification method itself: the original graph can be pruned such that the only remaining weighted edges are those that were identified by the successive MSTs, with the resulting graph containing $k(n-1)$ edges.

\subsection{Local Features}
The work of Sun et al. \cite{sun2020generalization} constructs four local features, comparing weights of an edge $(i,j) \in E$ to the edge weights $(k,j), k \in V$ and $(i,k),k\in V$. Analysis of these local features shows that they guide the decision-making process of classifiers most strongly. While relatively inexpensive to compute, yet less expensive features can be computed, comparing a given edge weight $(i,j)$ to the maximum and minimum weights in $E$. That is, for each $(i,j) \in E$ we compute a set of features $q_{ij}$ as:

\begin{equation}
    q^{a}_{ij} = (1 + A_{ij}) / (1 + \max_{(l,k)\in E}{A_{lk}})
    \label{a} 
\end{equation}
\vspace{-0.5em}
\begin{equation}
    q^{b}_{ij} = (1 + A_{ij}) / (1 + \max_{l\in V}{A_{il}})
    \label{b} 
\end{equation}
\vspace{-0.5em}
\begin{equation}
    q^{c}_{ij} = (1 + A_{ij}) / (1 + \max_{l\in V}{A_{lj}})
    \label{c} 
\end{equation}
\vspace{-0.5em}
\begin{equation}
    q^{d}_{ij} = (1 + \min_{(l,k)\in E}{A_{lk}}) / (1 + A_{ij})
    \label{d}
\end{equation}
\vspace{-0.5em}
\begin{equation}
    q^{e}_{ij} = (1 + \min_{l\in V}{A_{il}}) / (1 + A_{ij})
    \label{e}
\end{equation}
\vspace{-0.5em}
\begin{equation}
    q^{f}_{ij} = (1 + \min_{l\in V}{A_{lj}}) / (1 + A_{ij})
    \label{f}
\end{equation}

\noindent
The motivation for the features (\ref{a}) and (\ref{d}) is to cheaply compute features that relate a given edge weight to the weights of the entire graph in a global manner. On the other hand, motivated by the work of Sun et al. \cite{sun2020generalization}, the features (\ref{b}), (\ref{c}), (\ref{e}), (\ref{f}), compare a given edge weight to those in its direct neighbourhood; the edge weight associated with edge $(i,j)$ is related only to the weights of the associated vertices $i$ and $j$.

\subsection{Postprocessing Pruned TSP Graphs}
\begin{figure}[h]
    \centering
    \includegraphics[width=\textwidth]{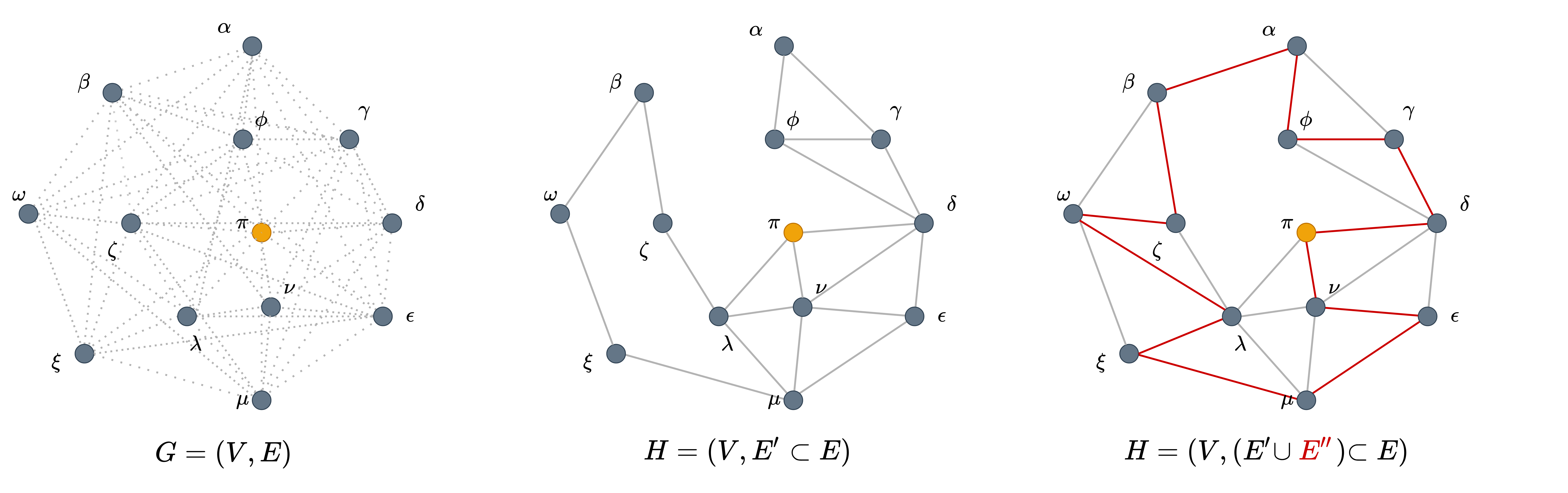}
    \caption{The pruned graph $H$ does not admit a feasible solution. \textit{Given a known solution, we add the edges $E''$ of the solution to the pruned graph in order to guarantee the feasibility of the pruned instance. If we obtain $E''$ using an approximation algorithm, then we can make guarantees about the quality of the solutions obtained from $H$.}}
    \label{fig:reconnecting}
\end{figure}

In this setting, sparsification is posed as a set of $m$ independent classification problems. The result of this is that there is no guarantee that any feasible solution exists within a pruned problem instance. Indeed, even checking that any tour exists within a sparsified graph is itself an NP-hard problem. One can guarantee feasibility of the pruned graph by ensuring that the edges belonging to some known solution exist in the pruned graph; this forces both connectivity and Hamiltonicity (see Figure \ref{fig:reconnecting}). The pruned graph has at least one feasible solution and admits an optimal objective no worse than that of the solution that is known. For the TSP we can construct feasible tours trivially by providing any permutation of the vertices, but it is likely that such tours will be far from optimal. Assuming the problem is metric, we can use an approximation algorithm to construct a feasible solution to the problem that also gives a bound on the quality of the solution.

\section{Experiments and Results}
\label{sec:experiments}
All experiments were carried out in Python\footnote{All code available at: https://github.com/JamesFitzpatrickMLLabs/optlearn}. Graph operations were performed using the NetworkX package \cite{hagberg2008exploring} and the training was carried out with the Scikit-Learn package \cite{pedregosa2011scikit}. The linear programming features were computed using the Python interfaces for the Xpress and SCIP  optimisation suites \cite{ashford2007mixed, maher2016pyscipopt}. Training and feature computation was performed on a Dell laptop running Ubuntu 18.04 with 15. 
6 GB of RAM an,  Intel® Core™ i7-9750H 2.60GHz CPU and an Nvidia  GeForce RTX 2060/PCIe/SSE2 GPU.

\subsection{Learning to Sparsify}
First we train a classification model to prune edges that are unlikely to belong to an optimal tour. That is, given the feature representation $$\vec{q}_{ij} = [ q^{a}_{ij}, q^{b}_{ij}, q^{c}_{ij}, q^{d}_{ij}, q^{e}_{ij}, q^{f}_{ij}, \hat{\vec{r}}_{ij}^{k},  \tilde{\vec{r}}_{ij}, \hat{q}^{k}_{ij}]^{T}$$ for the edge $(i, j)$, we aim to train a machine learning model that can classify all the edges of a given problem instance in this manner. In each case we let the parameter $k = \lceil \log_{2}(n) \rceil$, since this number grows slowly with the order $n$ of the graph and prevents excessive computation. In order to address the effects of class imbalance, we randomly undersample the negative class such that the classes are equally balanced, and adopt class weights $\{0.01, 0.99\}$ for the negative and positive classes respectively to favour low false negative rates for the positive class, favouring correctness over the sparsification rate. 

\hspace{1em}

\noindent
In many instances there exists more than one optimal solution. We compute all optimal solutions and assign unit value to the label for edges that belong to any optimal solution. Any optimal solution gives an upper bound on the solution for any subsequent attempts to solve the problem. To compute each solution, one can introduce a tour elimination constraint to the problem formulation to prevent previous solutions from being feasible. The problem can then be re-solved multiple times to obtain more solutions to the problem, until no solution can be found that has the same objective value as that of the original optimal tour.

\hspace{1em}

\noindent
Training was carried out on problem instances of the MATILDA problem set \cite{smith2010understanding}, since they are small ($n=100$ for each) and permit relatively cheap labelling operations. One third $(63)$ of the CLKhard and LKCChard problem instances were chosen for the training set, while the remaining problem instances are retained for the testing and validation sets. These two problem categories were selected for training once it was identified through experimentation that sparsifiers trained using these problems generally performed better (with fewer infeasibilities). This is in line with the findings of Sun et al.\cite{sun2020generalization}. Since each of the linear programming features must be computed for any given problem instance, it is worthwhile checking if any solution $\hat{\vec{x}}^{0}$ or $\tilde{\vec{x}}^{k}$ is an optimal TSP solution, which would allow all computation to terminate at this point.  All of the edges of a given graph belong to exactly one of training, test or validation sets, and they must all belong to the same set. Each symmetric problem instance in the TSPLIB problem set \cite{reinelt1991tsplib} for which the order $n \leq 4432$ is retained for evaluation of the learned sparsifier only. Sample weights $w_{ij} = A_{ij} / \max_{ij}A_{ij}$ are applied to each sample to encourage the classifier to reduce errors associated with longer edges.

\hspace{1em}

\noindent
Following pruning, edges of a known tour are inserted, if they are not already elements of the edge set of the sparsified graph. In this work, we compute both the double-tree and Christofides approximations
, since they can be constructed in polynomial time and guarantee that the pruned graph will give an optimality ratio $\hat{\ell}_{opt} / \ell_{opt} \leq 3/2$, where $\hat{\ell}_{opt}$ is the optimal tour length for a pruned problem and $\ell_{opt}$ is the optimal tour length for the original problem. The performance of the classifier is evaluated using the optimality ratio $\tilde{\ell} = \hat{\ell}_{opt} / \ell_{opt}$ and the retention rate $\tilde{m} = \hat{m} / m$, where $\hat{m}$ is the number of edges belonging to the pruned problem instance. Since the pruning rate $1 - \tilde{m}$ implies the same result as the retention rate, we discuss them interchangeably. Introducing any known tour guarantees feasibility at a cost in the sparsification rate no worse than $(\hat{m} + n) / \hat{m}$. 

\subsection{Performance on MATILDA Instances}

The sparsification scheme was first evaluated on the MATILDA problem instances, which are separated into seven classes, each with $190$ problem instances. Logistic regression, random forest and linear support vector classifiers were evaluated as classification models. Although no major advantage was displayed after a grid search over these models, here, for the sake of comparison with Sun et al., an L1-regularised SVM with an RBF kernel is trained as the sparsifier. We can see from Table \ref{tab:matildacomp} that the optimality ratios observed in the pruned problems are comparable to those of Sun et al, with a small deterioration in the optimality gap for the more difficult problems (around 0.2\%) when the same problem types were used for training. However, this comes with much greater sparsification rates, which leads to more than 85\% of the edges being sparsified in all cases and almost 90\% on average, as opposed to the around 78\% observed by Sun et al. We similarly observe that the more difficult problems are pruned to a slightly smaller extent than the easier problem instances.

\begin{table}[h!]
\begin{adjustbox}{center}
\begin{tabular}{|l||c|c|c|c|c|c|c|}
 \hline
 & \multicolumn{7}{|c|}{Problem Class} \\
\hline
Statistic (Mean) & CLKeasy & CLKhard & LKCCeasy & LKCChard & easyCLK-hardLKCC & hardCLK-easyLKCC & random \\
\hline
Optimality Ratio & 1.00000 & 1.00176 & 1.00000 & 1.00385 & 1.00049 & 1.00035 & 1.00032 \\
\hline
Pruning Rate & 0.90389 & 0.89021 & 0.91086 & 0.88298 & 0.88874 & 0.90357 & 0.89444\\
\hline  
\end{tabular}
\end{adjustbox}

\caption{Evaluation of the trained sparsifier against the problem subsets of the MATILDA instances. Here each cell indicates the mean value of the optimality ratio or pruning rate over each subset of problems (not including training instances)}\label{tab:matildacomp}
\end{table}

\subsection{Pruning With and Without Guarantees}

\begin{figure}[h!]
    \centering
    \includegraphics[width=0.9\textwidth]{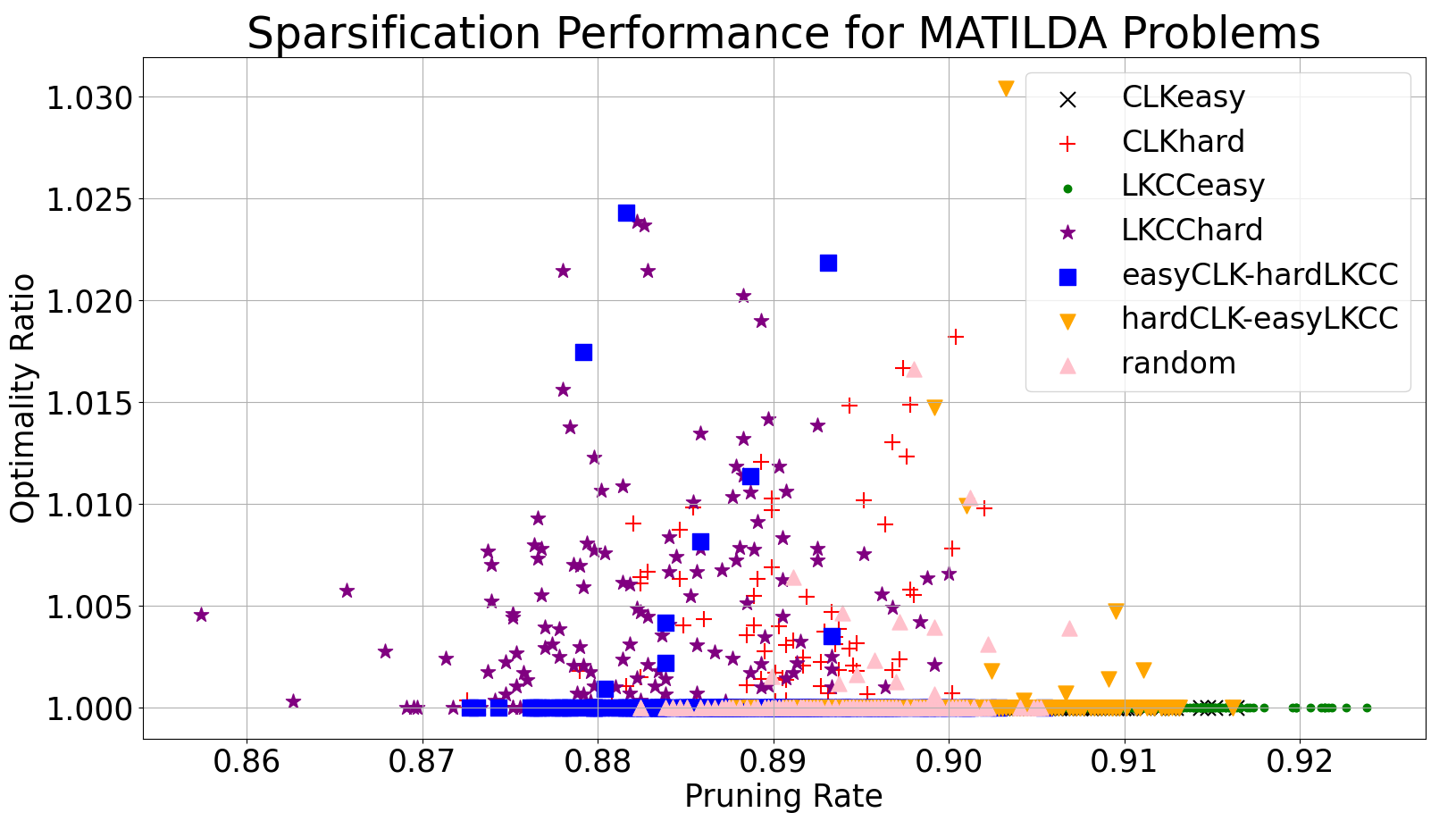}
    \caption{Each point represents a problem instance, showing the pruning rate and the optimality ratio achieved for each problem in the MATILDA benchmark set.}
    \label{fig:matplot}
\end{figure}

This section describes experiments carried out to test how much the sparsifier would need to rely on inserted tours to produce feasibly reduced problem instances. The optimality ratio statistics are shown in Table \ref{tab:matildares}. Before post-processing, pruning was achieved with a maximum rate of edge retention $\max\{\hat{m}/m\} = 0.179$, a minimum rate $\min\{\hat{m}/m\}=0.140$ and a mean rate $\langle \hat{m}/m \rangle = 0.143$ among all problem instances. In the vast majority of cases, at least one optimal tour is contained within the pruned instance (column 2), or one that is within $5$\% of optimal (column 4). For the CLKhard problem instances, there were one to three infeasible instances produced each time. Only in the case of the random problem instances did the number of pruned problems containing optimal solutions increase after inserting the approximate tour edges (column 2). In some cases (for easyCLK-hardLKCC, CLKhard and random) the pruned instances admitting sub-optimal solutions with respect to the original problem had improved optimality ratios (columns 3 and 4). The worst-case optimality ratios changes little for each problem class  after post-processing, except for the case (CLKhard) where previously infeasibile pruned instances made it impossible to compute $\tilde{\ell}$ (column 5). The distribution of these values with post-processing is depicted in Figure \ref{fig:matplot}.

\begin{table}[h!]
\centering
\begin{tabular}{|l||x{6.5em}|c|c|c|c|}
 \hline
 & \multicolumn{3}{|c|}{\# of Problems With Below Condition True} & Worst Case \\
\hline
Problem Class     & $\hat{\ell}_{opt} / \ell_{opt} = 1.0 $  & $\hat{\ell}_{opt} / \ell_{opt} < 1.02 $    & $\hat{\ell}_{opt} / \ell_{opt} < 1.05 $ & $\max\{\hat{\ell}_{opt} / \ell_{opt}\}$ \\
\hline
\hline
CLKeasy             &   $190 \rightarrow 190$     &   $190 \rightarrow 190$     &       $190\rightarrow190$        &        $1\rightarrow1$        \\
\hline  
CLKhard             &    122 $\rightarrow122$    &   182 $\rightarrow188$     &       $185\rightarrow190$        &        $\infty\rightarrow1.024$        \\
\hline
LKCCeasy             &  $190\rightarrow190$      &  $190\rightarrow190$      &      $190\rightarrow190$         &       $1\rightarrow1$         \\
\hline
LKCChard            &   $89\rightarrow89$     &   $184\rightarrow184$     &    $189\rightarrow189$           &         $1.054\rightarrow1.054$       \\
\hline
easyCLK-hardLKCC             &    $180\rightarrow180$    &  $185\rightarrow190$      &      $190\rightarrow190$         &       $1.010\rightarrow1.010$         \\
\hline  
hardCLK-easyLKCC             &   $184\rightarrow184$     &    $190\rightarrow190$    &        $190\rightarrow190$       &        $1.015\rightarrow1.015$        \\
\hline  
random             &     $174\rightarrow179$  &    $189\rightarrow190$    &       $190\rightarrow190$        &       $1.046\rightarrow1.0043$         \\
\hline
\hline
Total             &   $1129\rightarrow1134$     &   $1310\rightarrow1324$     &       $1324\rightarrow1329$        &         $\infty\rightarrow1.054$       \\
\hline  

\end{tabular}

\caption{Optimality ratio statistics before and after approximate tour insertion following learned sparsification (MATILDA problem instances). \textit{Each cell indicates the number of problem instances of each class for which the optimality ratio resided within the bounds stated before and after post-processing. For example, the cell in column 2 containing $174\rightarrow179$ indicates that $174$ purely sparsified instances admitted unit optimality ratios, but $179$ did so after the post-processing. The last column contains the maximum optimality gap for each class. If there are infeasible sparsified graphs, this value is denoted by $\infty$.} }\label{tab:matildares}
\end{table}

\noindent
This sparsification scheme was also tested on the TSPLIB problem instances (see Table \ref{tab:tsplibres}). Before post-processing, for the majority of the problem instances, an optimal tour is contained within the sparsified graphs (column 2), and the vast majority of the problem instances have optimality ratios no worse than $\tilde{\ell} = 1.05$ (column 4). Unlike with the MATILDA instances, the pruning rate varies significantly. This is in accordance with the findings of Sun et al.\cite{sun2020generalization} and indicates that the sparsifier is less certain about predictions made, and fewer edges are therefore removed in many cases. For some smaller problem instances (smaller than the training set problems) the pruning rate approaches $0.5167$. The median pruning rate was $0.9109$, with the mean pruning rate at $0.8904$ and a standard deviation of $0.086$. The highest pruning rate was for the problem \textit{d657.tsp}, for which $0.9618m$ edges were removed, achieving an optimality ratio of $1.0$. Three instances were found to be infeasible under this scheme without approximation, \textit{si535.tsp}, \textit{pr299.tsp}, and \textit{pr264.tsp}.

\hspace{1em}

\begin{figure}[h!]
    \centering
    \includegraphics[width=0.9\textwidth]{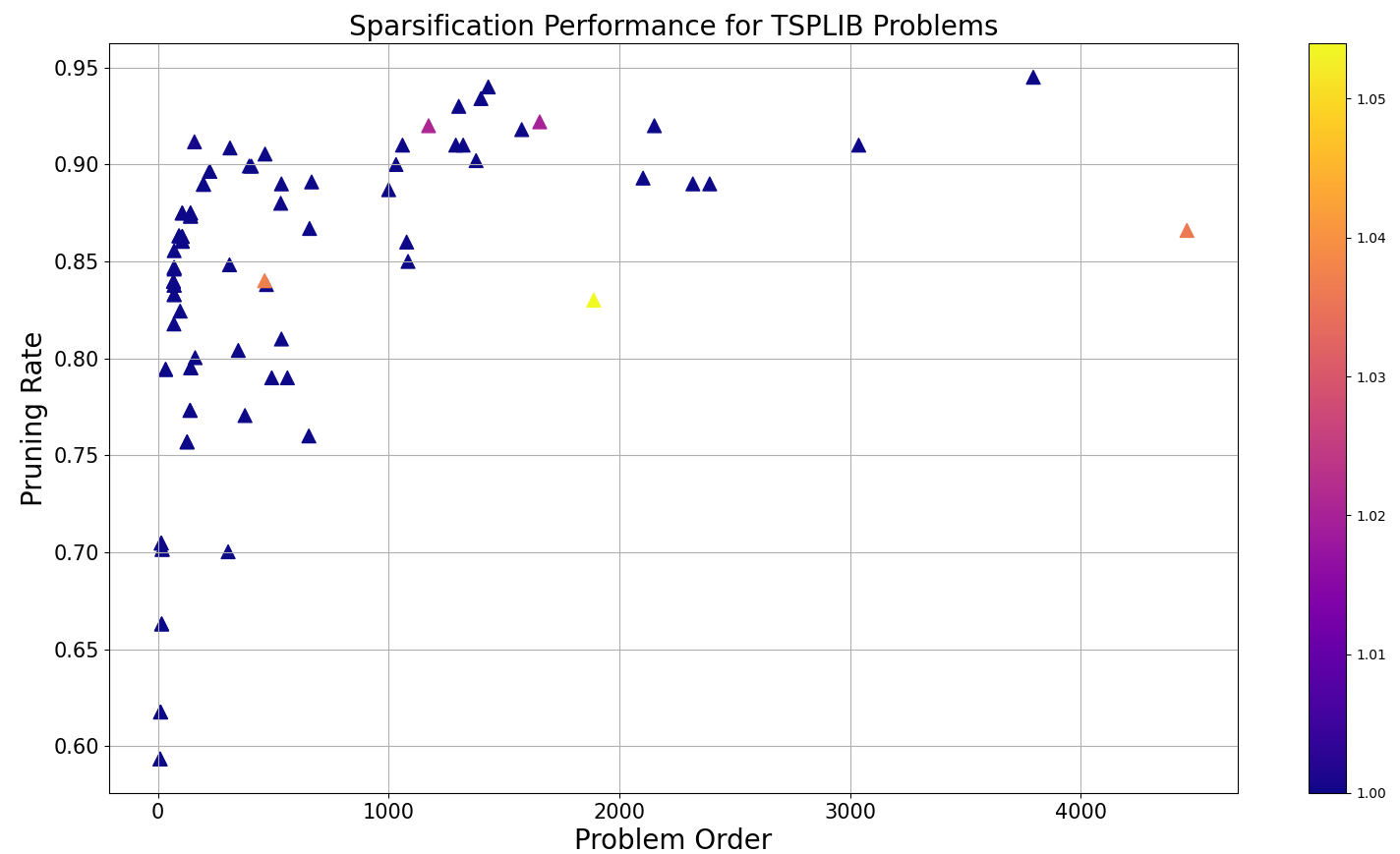}
    \caption{Here each point represents a problem instance. The horizontal axis depicts the problem size in terms of the number of nodes. The vertical axis shows the sparsification rate and the colouring of the points indicates the optimality ratio observed.}
    \label{fig:tsplibplot}
\end{figure}

\noindent
Introducing the approximate tours brought the worst-case optimality ratio to $1.01186$ (column 5) for \textit{pr264.tsp}, with median and mean pruning rates of, respectively, $0.8991$ and $0.8718$. The mean optimality ratio (excluding the infeasible instances) before double-tree insertion was $1.00092$ and (also without these same problems) $1.00073$ after insertion, indicating that in most cases there is little reduction in solution quality as a result of sparsification. In Figure \ref{fig:tsplibplot} we can see depicted the relationship between the problem size (the order, $n$) and the pruning rate. Almost all of the instances retain optimal solutions after pruning, those that don't are indicated by the colour scale of the points.

\begin{table}[h!]
\centering
\begin{tabular}{|l||c|c|c|c|c|}
\hline
 & \multicolumn{3}{|c|}{\# of Problems With Below Condition True} & Worst Case \\
\hline
Problem Class     & $\hat{\ell}_{opt} / \ell_{opt} = 1.0 $  & $\hat{\ell}_{opt} / \ell_{opt} < 1.005 $    & $\hat{\ell}_{opt} / \ell_{opt} < 1.010 $ & $\max\{\hat{\ell}_{opt} / \ell_{opt}\}$ \\
\hline
\hline
TSPLIB             &   $55 \rightarrow 56$     &   $69\rightarrow70$     &      $72\rightarrow73$         &        $\infty\rightarrow1.01186$        \\
\hline  
\end{tabular}
\caption{Optimality ratio statistics before and after approximate tour insertion following learned sparsification (TSPLIB problem instances).  \textit{Results for each symmetric problem instance with $n \leq 1000$ using a logistic regression sparsifier. This set contained $76$ problem instances, some of which are not metric TSPs. Although we cannot make provable guarantees for the non-metric problem instances, we can still use the approximate tours to ensure feasibility.}}\label{tab:tsplibres}.
\end{table}

\subsection{Minimum Weight Spanning Tree Pruning}

Empirical experiments demonstrated that using only successive MSTs, as outlined in Section \ref{ssec:num1},  one can effectively sparsify symmetric TSP instances. This scheme proceeds without the need for a classifier, by building a new graph $H$ with only the edges of the MSTs and their associated edge weights in $G$. This sparsified instance achieves a retention rate of $kn/m$, where $k$ is the number of trees computed. In many cases, simply using this as a scheme for selecting edges for retention was sufficient to sparsify the graph while achieving a unit optimality ratio. 

\hspace{1em}

\noindent
On the MATILDA problems, again with $k = \lceil\log_{2}(n)\rceil$, thhis achieves a worst-case optimality ratio of $1.0713$ and a uniform pruning rate of $0.86$, with the majority of the pruned instances containing an optimal solution. Under these conditions the easier problems (CLKeasy and LKCCeasy) had optimal tours preserved in every problem instance except for one. For the more difficult problem instances, in the vast majority of cases, the ratio does not exceed $1.02$ (see Table \ref{tab:matildamwstres}). Insertion of approximate tours in this scenario does not lead to much improvement, since every instance of the MATILDA problem set is sparsified feasibly in this scheme.
\begin{table}[h!]
\centering
\begin{tabular}{|l||c|c|c|c|c|}
\hline
 & \multicolumn{3}{|c|}{\# of Problems With Below Condition True} & Worst Case \\
\hline
Problem Class     & $\hat{\ell}_{opt} / \ell_{opt} = 1.0 $  & $\hat{\ell}_{opt} / \ell_{opt} < 1.02 $    & $\hat{\ell}_{opt} / \ell_{opt} < 1.05 $ & $\max\{\hat{\ell}_{opt} / \ell_{opt}\}$ \\
\hline
\hline
CLKeasy             &   $190\rightarrow190$     &   $190\rightarrow190$     &       $190\rightarrow190$        &        $1\rightarrow1$        \\
\hline  
CLKhard             &    $85\rightarrow88$        &   $185\rightarrow185$     &       $190\rightarrow190$        &        $1.029\rightarrow1.029$        \\
\hline
LKCCeasy             &  $189\rightarrow190$      &  $190\rightarrow190$      &      $190\rightarrow190$         &       $1.002\rightarrow1.002$         \\
\hline
LKCChard            &   $77\rightarrow78$     &   $183\rightarrow183$     &    $189\rightarrow189$           &         $1.071\rightarrow1.054$       \\
\hline
easyCLK-hardLKCC             &    $173\rightarrow190$    &  $190\rightarrow190$      &      $190\rightarrow190$         &       $1.012\rightarrow1.012$         \\
\hline  
hardCLK-easyLKCC             &   $171\rightarrow173$     &    $190\rightarrow190$    &        $190\rightarrow190$       &        $1.015\rightarrow1.015$        \\
\hline  
random             &     $176\rightarrow176$   &    $190\rightarrow190$    &       $190\rightarrow190$        &       $1.006\rightarrow1.006$         \\
\hline
\hline
All             &    $1061\rightarrow1330$     &    $1318\rightarrow1330$     &      $1329\rightarrow1330$         &         $1.071\rightarrow1.054$       \\
\hline  
\end{tabular}
\caption{Optimality ratio statistics before and after approximate tour insertion following MST sparsification (MATILDA problem instances). \textit{Comparison between the pure multiple MST pruning scheme and the same scheme with double-tree post-processing. In every case for the MATILDA problem set, the pruned instances are feasible.}}\label{tab:matildamwstres}.
\end{table}

\vspace{-2.5em}

\noindent
For the TSPLIB problems, all but two can be sparsified feasibly without the insertion of approximate  edges: \textit{p654.tsp} and \textit{fl417.tsp}. Since $k$ is a function of the problem order, $n=100$, $86\%$ of the edges of a graph will be removed, whereas at order $n=1000$, the pruning rate reaches $98\%$. The worst optimality ratio for any feasible problem was for $pr226.tsp$, at $\tilde{l}=1.106$, with a retention rate of $0.071$. The majority of the sparsified instances retained an optimal tour (see Table \ref{tab:tsplibresmwst}), with the mean optimality ratio $\langle \tilde{\ell} \rangle$ for all feasible problems taking the value $1.0061$. Including approximate  tour edges in the sparsified graphs results in none having an optimality gap greater than $1.084$ (\textit{pr654.tsp}, which was previously infeasible, with a retention rate of $3.1\%$). The other previously infeasibly-sparsified problem, \textit{fl417.tsp} admits an optimality ratio $\tilde{\ell}=1.077$, retaining just $4.3\%$ of its edges. The mean optimality ratio for all problems emerged as $1.0053$.

\begin{table}[h!]
\centering
\begin{tabular}{|l||c|c|c|c|c|}
 \hline
 & \multicolumn{3}{|c|}{\# of Problems With Below Condition True} & Worst Case \\
\hline
Problem Class     & $\hat{\ell}_{opt} / \ell_{opt} = 1.0 $  & $\hat{\ell}_{opt} / \ell_{opt} < 1.02 $    & $\hat{\ell}_{opt} / \ell_{opt} < 1.05 $ & $\max\{\hat{\ell}_{opt} / \ell_{opt}\}$ \\
\hline
\hline
TSPLIB             &   $51 \rightarrow 51$     &   $66 \rightarrow 67$     &      $71 \rightarrow 71$         &        $ \infty \rightarrow 1.084$        \\
\hline  
\end{tabular}
\caption{Optimality ratio statistics before and after approximate tour insertion following MST sparsification (TSPLIB problem instances). \textit{For each problem instance the number of trees $k$ differed, according to the order $n$.}}\label{tab:tsplibresmwst}.
\end{table}

\subsection{Specifying the Pruning Rate}
One of the advantages of including the post-processing step that guarantees the feasibility of the pruned instance is that we can exert greater control over the pruning rate. Training a sparsifier in the manner outlined above necessitates a trade-off between the optimality ratio and the pruning rate. Higher pruning rates result in sparsified problems that are easier to solve but that have typically poorer optimality ratios. Indeed, if the sparsification rate is too high, many of the sparsified instances will also become infeasible. Including approximate tours, however, allows us to choose effectively any decision threshold for the classifier, ranging from total sparsification and removing all edges, or choosing a threshold that results in no sparsification at all and the retention of all edges from the original graph. So long as there is at least one known feasible tour, regardless of its quality, then the pruned instance will be feasible after post-processing.

\section{Discussion and Conclusions}
\label{sec:discussion}
In this work we have demonstrated that it is possible to learn to effectively sparsify TSP instances, pruning the majority of the edges using a small classification model while relying on linear programming and graph-theoretic features that can be efficiently computed. Providing guarantees of feasibility is possible by means of inserting edges belonging to approximate tours, and this ensures that even out-of-distribution problem instances can be tackled with this scheme. These features are supplemented with local statistical features, comparing edge weights to the global and neighbouring distributions. This scheme successfully generalises to larger problem instances outside of the training distribution. Where there is a lack of training data or where expediency is favoured over improvements that can be obtained from training a sparsifier, it has been shown that the MST extraction mechanism with inserted doubletours performs well on most problem instances. 

\hspace{1em}

\noindent
The motivation of this work has been to use the methods of ML to aid in the design of heuristics for solving combinatorial optimisation problems. This is in the hope that such an approach can reduce the development time required. Development time for ML solutions depends typically on the engineering of either features or problem-specific models and architectures. This can effectively transfer the the engineering effort from one task to another, without producing tangible benefits. In this approach, classical ML models are used, which means that feature design is paramount to the success of the model. Here, LP features are designed that are not dependent upon the formulation of the problem; any TSP problem that requires formulation as binary IP with subtour-elimination (or capacity) constraints can have features produced in the same manner. This is attractive because it simply requires knowledge of IP formulations that give tight relaxations. Such relaxations may also be sufficient to solve the desired problem to begin with, potentially obviating the need for any further computation. Optimality ratio performance does not appear to depend on the problem size, which indicates that this pre-processing scheme might help extend the applicability of learned solving heuristics.

\hspace{1em}

\noindent
The use of the double-tree approximation guarantees the feasibility of the pruned problem instance, but it does not provide a tight bound on the optimality ratio. Tighter bounds can be achieved by using stronger approximation algorithms, such as that of Christofides and Serdyukov \cite{christofides1976worst, serdyukov1978}, at a cost of greater running times. Alternatively, one could use multiple approximation algorithms, in order to increase the number of tours inserted into the pruned instance, or compute successive approximations by removing from consideration edges belonging to a tour once they have been computed, analogously to Algorithm \ref{algo_mwst}. The results obtained indicate that, in the case of the double-tree, it did not improve the worst-case performance in general, but did improve the mean performance in terms of the optimality ratio. This suggests that although the additional edges do not lead to significant improvements, it is not necessary in most cases to rely on them to obtain acceptable optimality ratios, just to guarantee feasibility. Therefore, insertion of any feasible tour may be sufficient for routing problems where no effective or efficient approximation scheme exists.

\hspace{1em}

\noindent
This scheme has the potential to be developed in a similar manner for other routing problems, in particular vehicle routing problems, for which solvers are not as effective in practice as they are for the TSP. To realise the benefits of scheme, an implementation would also have to be rewritten in a lower level language. Subsequent work could be done to evaluate the smallest problem sizes for which training can be carried out effectively and applicability to the VRP, for which sparsification is already a natural process if there are time windows.

\section*{Acknowledgements}

This work was funded by Science Foundation Ireland through the SFI Centre for Research Training in Machine Learning (18/CRT/6183).

\bibliographystyle{plain} 
\bibliography{references}


\end{document}